# Using LLM such as ChatGPT for Designing and Implementing a RISC Processor: Execution, Challenges and Limitations

Shadeeb Hossain [ORCID ID: 0000-0002-5224-7684], Aayush Gohil, Yizhou Wang

*Abstract*—**This paper discusses the feasibility of using Large Language Models (LLM) for code generation with a particular application in designing an RISC. The paper also reviews the associated steps such as parsing, tokenization, encoding, attention mechanism, sampling the tokens and iterations (optional) during code generation. The generated code for the RISC components is verified through testbenches and hardware implementation on a FPGA board. Four metric parameters: (i) Correct output on the first iteration (ii) Number of errors embedded in the code (iii) Number of trials required to achieve the code and (iv) Failure to generate the code after three iterations ; are used to compare the efficiency of using LLM in programming. In all the cases, the generated code had significant errors and human intervention was always required to fix the bugs. LLM can therefore be used to complement a programmer's code design.**

***Index Terms***— ***ChatGPT, Code generation, FPGA, LLM, Programming*** .

## I. INTRODUCTION

LARGE Language Models (LLM) can comprehend the general-purpose language and produce outputs via training through large quantity of resources and are complemented with supervised learning [1-4]. LLM has also been in code generation and various prompts can be used to get the desired outputs [5,6]. GitHub repositories or other similar code-based data are used for training these LLM and have been successful at generating code-based outputs to perform certain subsets of tasks [7-9]. There has been an increasing trend to train models on large scale code corpora and can perform the relevant code generation task without requiring expensive fine tuning [10,11].

Sematic parsing maps which include converting a natural language (NL) utterance to a machine executable logic has been used in miscellaneous code generation tasks [12]. It is a two-step process: (i) *Step-1*: predicting a preliminary sketch (code structure) which ignores the low-level details. (ii) *Step-II* : fill in the details from analyzing the NL and the above generated sketch.

Another encoder-decoder architecture that synthesizes visualization programs is PLOTCODER [13]. It is a deep neural network code generation model which also generates the code from a combination of NL utterances and code context. The model processes the following steps: (i) Analyzing the NL description (ii) *Local code context* :it includes a few initial lines of code (iii) *Distant data frame code*: it includes the data frame manipulation (iv) *Data frame schema* and (v) *Ground truth*.

The neural network architecture used by ChatGPT 3.5 (Generative Pre-Trained Transformer) is a transformer model [14]. It has an encoder-decoder structure where the encoder maps the input to a continuous representation and then generates an output sequence using the decoder. The encoder consists of 6 layers and each layer has 2 sublayers : (i) *multi-head self-attention mechanism* (ii) *position wise fully connected feed-forward network.* However, the decoder also has 6 layers but has 3 sublayers: (i) *multi-head self-attention mechanism* (ii) *position wise fully connected feed-forward network* and (iii) *the third sublayer performs multi-head attention over the output of the encoder stack.*

This paper focuses on using LLM such as ChatGPT 3.5 in generating and implementing the VHDL code for a RISC. RISC is an open-source Instruction Set Architecture (ISA) that has the potential to improve microprocessor design, reduce computational cost and ease the transition to specialized tasks [15]. The component in our design focuses on the following: (i) Program Counter (PC) (ii) Register File (iii) Arithmetic Logic Unit ( ALU) (iv) Control Unit (CU) (v) Data Memory (vi) Instruction Memory. The details of each of the components and its function are discussed in detail in the following section.

LLM was used to generate the code for each component and the corresponding testbenches were also generated using ChatGPT 3.5 for verification. VHDL (VHSIC Hardware Description Language) was used as our programming language and the design was implemented and tested using BASYS 3

Shadeeb Hossain is with Department of Electrical and Computer Engineering, New York University (e-mail: sh7492@nyu.edu ).
Aayush Gohil is with Department of Electrical and Computer Engineering, New York University (e-mail: apg9124@nyu.edu).
Yizhou Wang is with Department of Electrical and Computer Engineering, New York University (e-mail: yw7818@nyu.edu).

FPGA board. The paper is divided into the following sections : (i) Background about using LLM in the RISC architecture – this discusses the steps associated with 'parsing', 'tokenization, 'attention mechanism' and other relevant steps used by LLM to generate the desired code. It also discusses the metric used to test the efficiency of each code. (ii) Background about RISC Architecture: It briefly describes the design path and the architecture of the RISC. (iii) Results and Discussion – includes individual testbenches and combined high level testbenches. It also discusses the challenges associated in using LLM in programming for each of the components.(iv) Conclusion- It includes a summary and discusses the challenges of relying solely on LLM for programming .

## II. Background about using LLM in RISC Architecture

Fig.1 shows the example of three prompts that have been used to generate the VHDL code for Program Counter (PC) , Register file and Arithmetic Logic Unit (ALU). The prompts play a critical role in generating the desired outputs because any missing information might ignore an essential feature code critical to implementing the processor. When a command is prompted to VHDL the following steps take place as shown in Fig.2 (a). The general steps associated with LLM for code generation have been discussed in several papers that focuses on ChatGPT Transformer models, tokenization, and decoding strategies [14, 16-19].

The initial step is "*Parsing*" which includes extracting and analyzing the prompts to find the relevant information. This key information needs to be analyzed to generate the relevant code. For example, in context of ALU, the LLM should be able to draw relevant relation between the operation of the ALU such as addition, subtraction , comparison and its corresponding (i) control signals , (ii) sources of the registers, etc. The code is then generated using GPT (Generative Pre-Trained Transformer). The associated steps of the code generation are discussed in Fig. 2b. The generated code is ensured to follow relevant guidelines and standards. This is to ensure that all the industry standards are optimized. It is important to understand that iterative refinement ( is optional) depending on the initial feedback on running the VHDL code.

As shown in Fig. 2(b), the first step in code generation is tokenization into words or sub words. The punctuations are also treated as separate entities and therefore assigned a token [14,16] . It would involve methods like byte pair encoding (BPE). The tokens are associated with an embedded vector. The LLM models process the input sequences from left to right and generate a context-based representation from input tokens.

In transformer model, the attention mechanism is applied to "selectively focus" on the input sequence of tokens [14,20]. In the self-attention mechanism "attention scores" are assigned for each token based on the relationship with their peer tokens. This is critical as it helps to determine the attention each token should attain. The "attention scores" also helps to evaluate the calculated weighted sum. It plays a pivotal role in capturing the dependencies and context. In sampling tokens: the tokens are sampled and each previously generated token along with the input context is used to generate the next token.

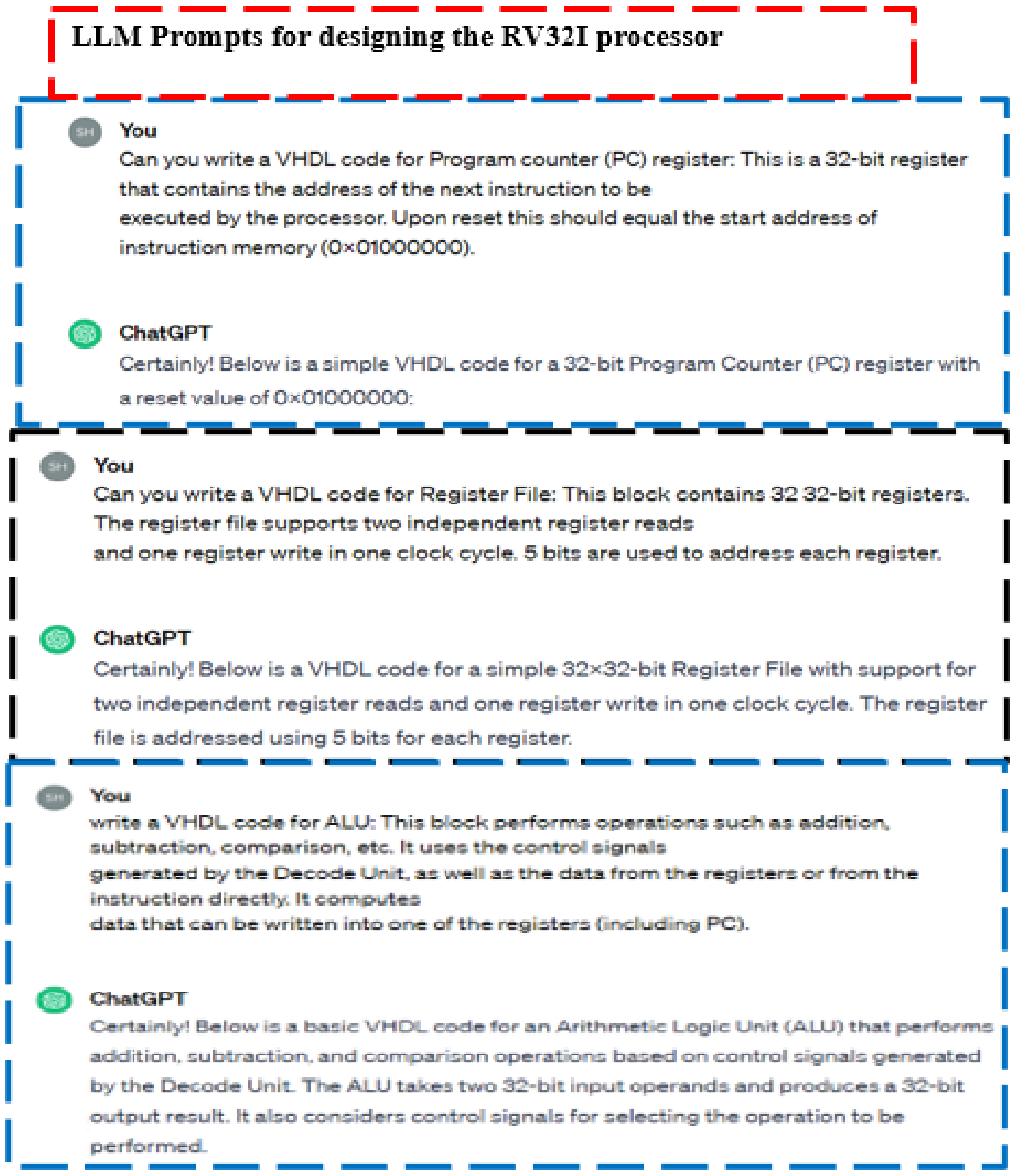

**Fig. 1 :** The ChatGPT prompts to generate the VHDL codes for (i) Program Counter (PC) (ii) Register file (iii) Arithmetic Logic Unit (ALU)

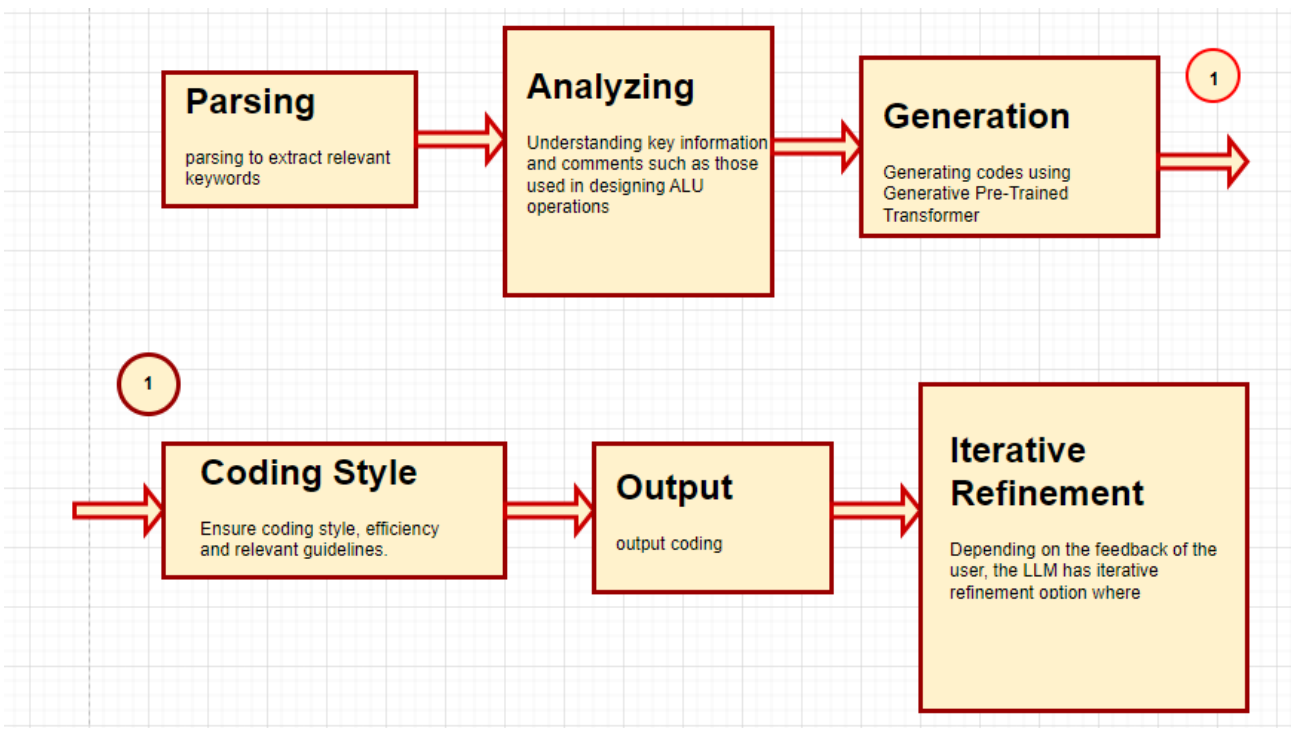

**Fig.2 (a) :** Sequence of general steps for VHDL code generation (for example in programming for ALU components) using LLM

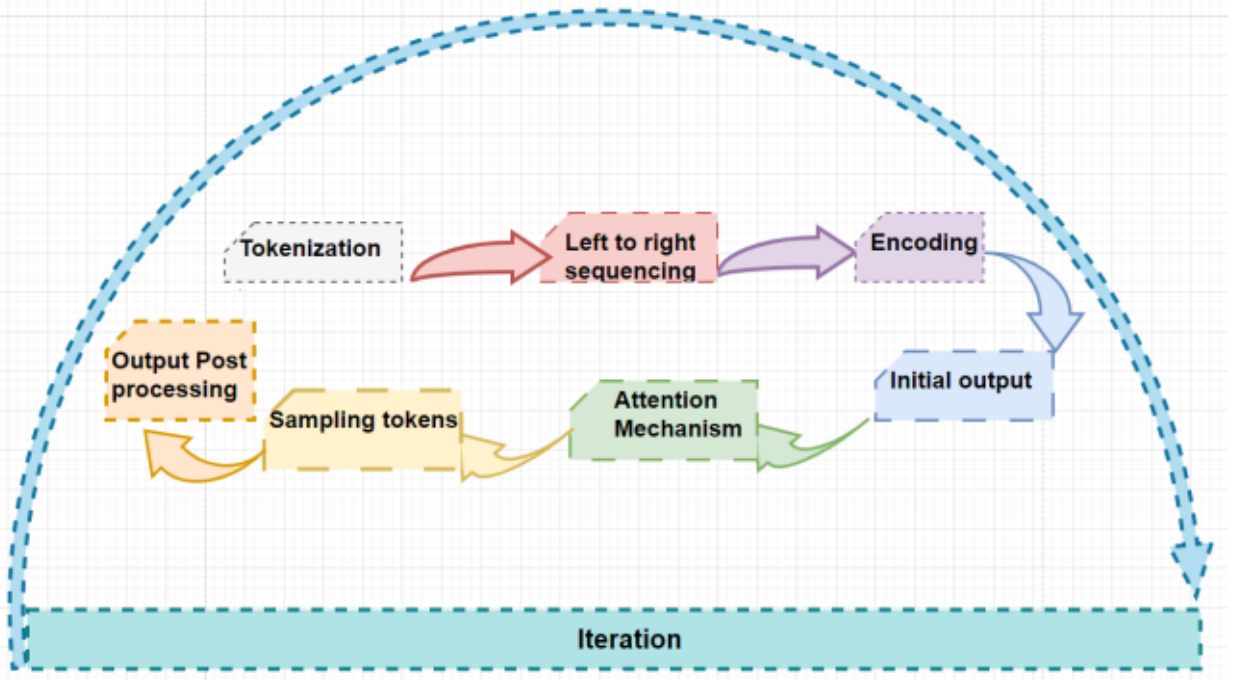

**Fig. 2(b)** : Steps involved in the Generative Pre-Trained Transformer section of Code generation for VHDL.

To test the efficiency of using ChatGPT LLM model, we assigned the following metric rubric as shown in Fig. 3. The generated code is evaluated based on the following components : (i) Correct output on initial iteration ( binary classification can be used as ‘0’: *failed to be correct on first iteration* and ‘1’ : *correct on first iteration* (ii) Number of errors embedded in the code (iii) Number of trials required to achieve the correct code and (iv) Failure to generate the correct code after three iterations (this can also be a binary classification as *‘0’ : if not failed* or *‘1’: if failed to generate the correct code).*

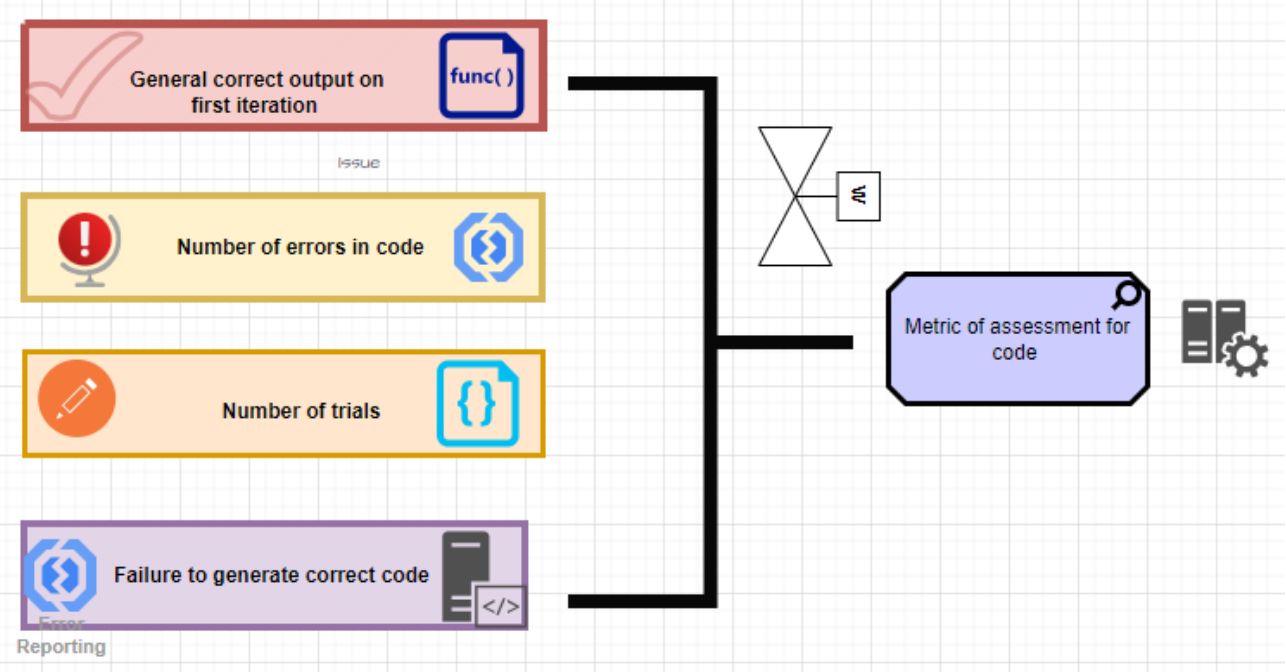

**Fig. 3** : Metric parameters used to assess the efficiency of the VHDL code generated from ChatGPT.

## II. Background about RISC Architecture

RV321 is an instruction set architecture (ISA) that contains 40 unique instructions [15, 21-23]. It is designed to form a compiler target and support modern operating systems (OS) environment.

The architecture of the RV321 was kept simple with focus on the following core components discussed below:

### *A. Program Counter (PC)*

The program counter for the RISC is a 32-bit register and its function is to point to the next instruction. There is to be a reset pin that should allow the counter to set to its default value which is *“01000000000000000000000000000000”*.A VHDL code is generated using the flowchart as shown in Fig. 4, which focuses on the operation of the PC discussed above . Vivado Design Suite was used to synthesize and analyze all the VHDL code with their testbenches.

### *B. Register File*

The Register File for the RV32I processor is a crucial component that stores the data to be processed and the results of various computations. It consists of a set of registers, each of which is a 32-bit storage unit. Fig. 5 shows the operation of the register files and the programming performed using VHDL code.

### *C. Control Unit:*

The control unit generates the proper control signal in response to the 32-bit instructions in an “R-format”. The type of instruction then determines which control signals are to be asserted and what function the ALU, and other components needs to perform.

Fig. 6 shows the flowchart for a Control Unit architecture . The process is broken down into two steps:

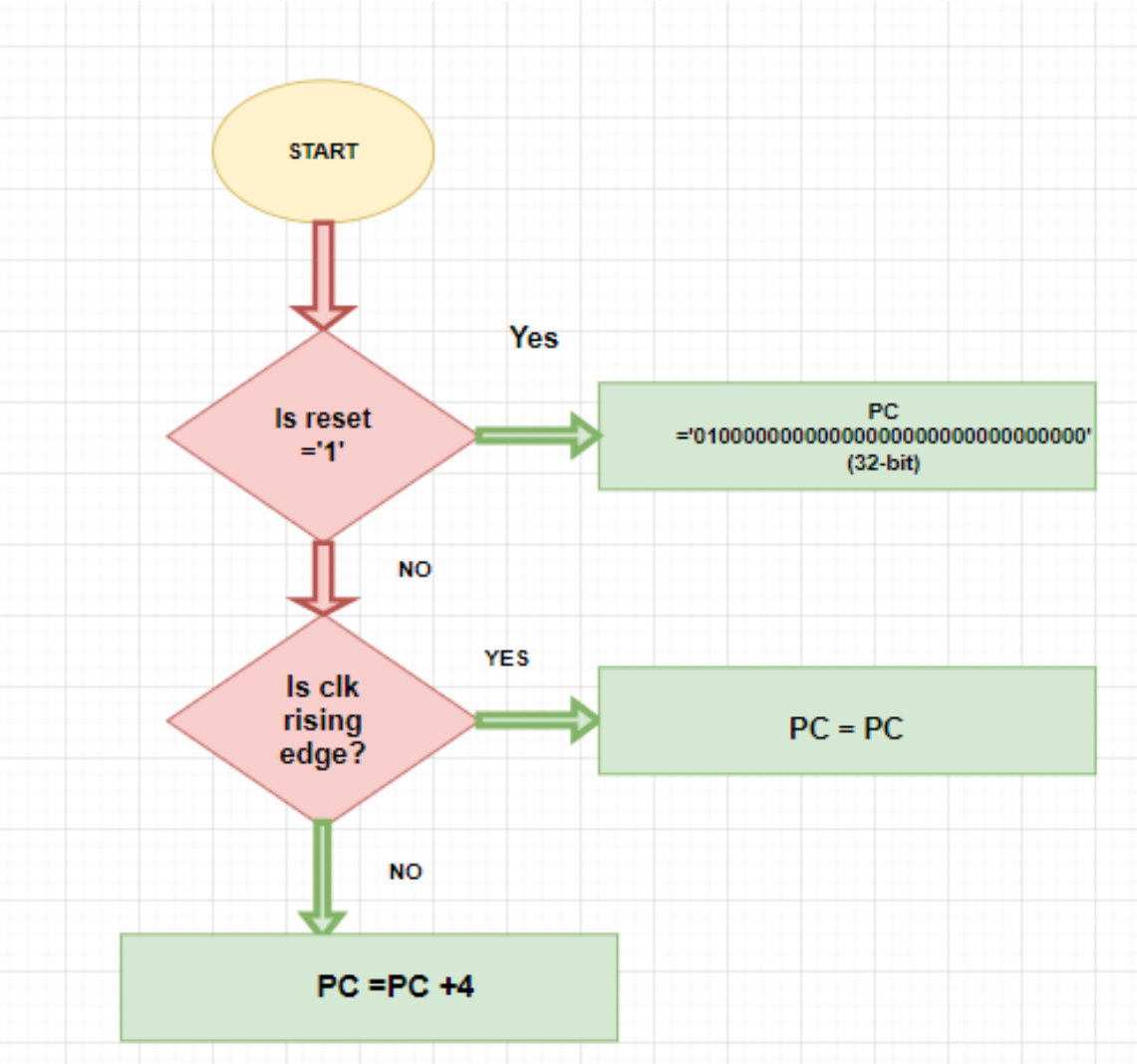

**Fig. 4** : Flowchart showing the operation of Program Counter (PC).

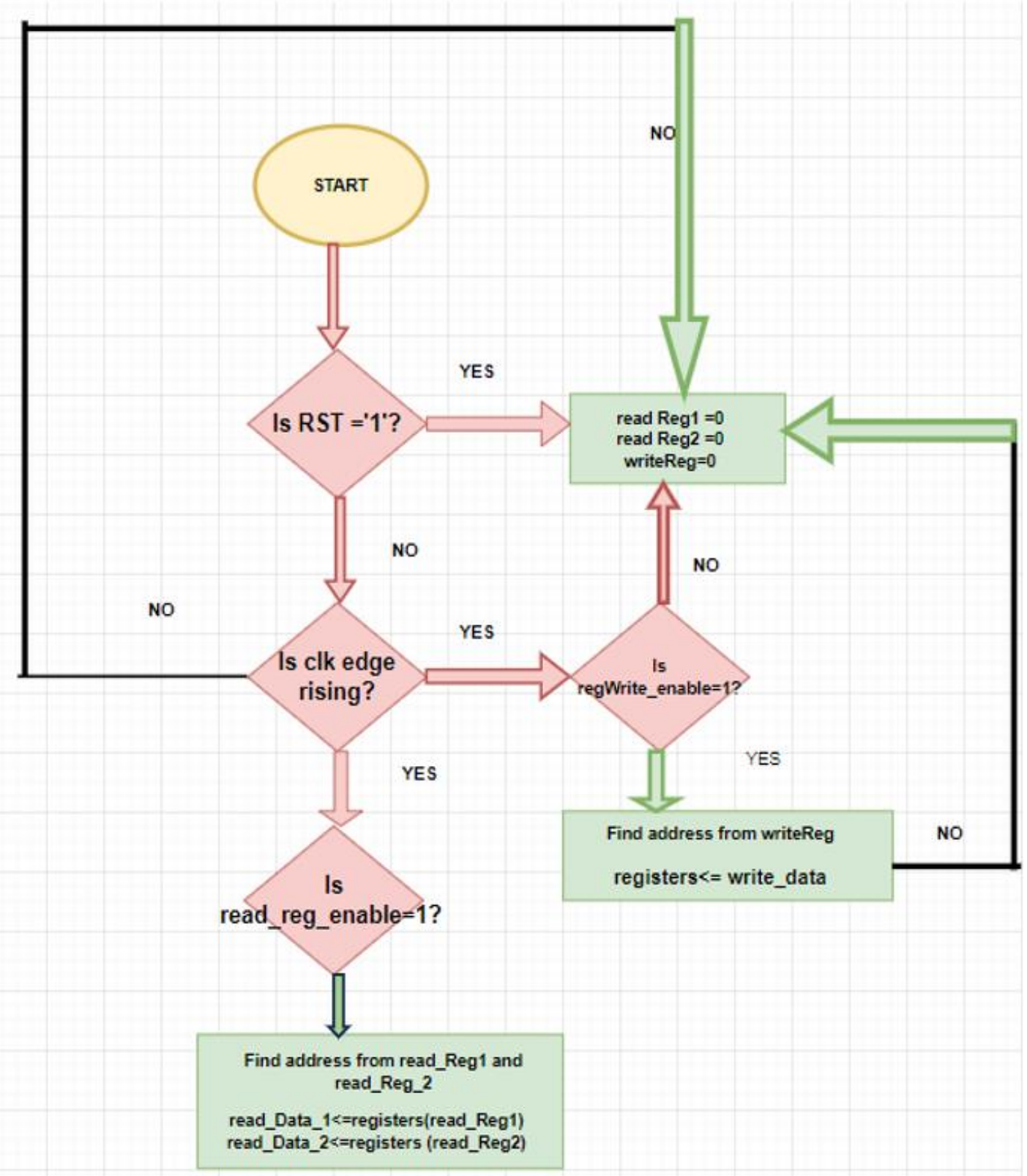

**Fig. 5** : Flowchart showing the operation of Register files

For this flowchart, we only focused on LUI ( Load Upper Immediate), however during implementation of the code all the 40 instructions were implemented.

### D. *Arithmetic Logic Unit (ALU)*

It performs operations such as addition, subtraction, and comparison. The instruction set is used to generate the relevant ALU . The component uses the control signal generated by the ‘decode unit’ to operate.

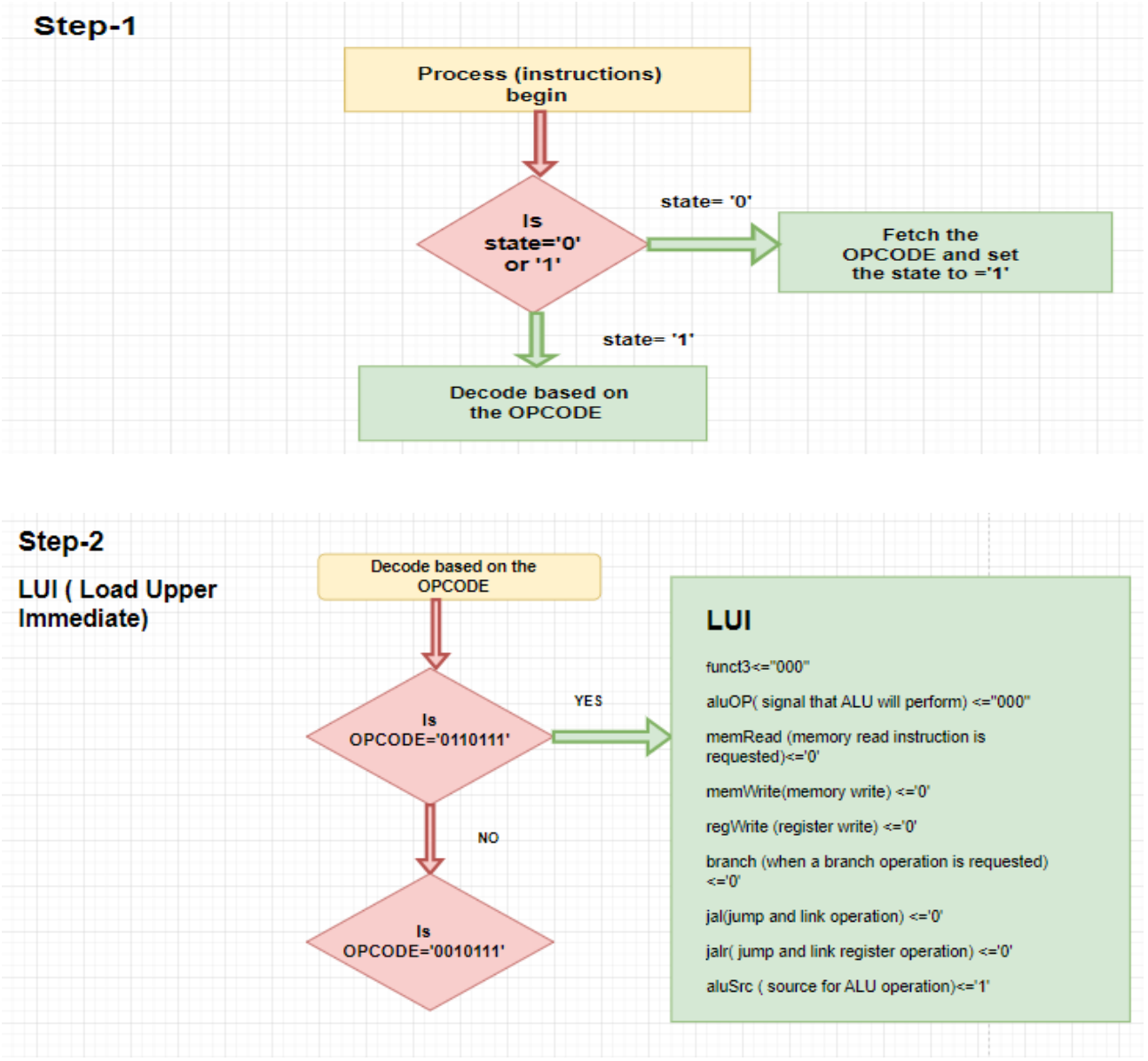

**Fig. 6**: Flowchart showing the operation of a Control Unit in a RISC. Step-1 includes *fetching* the OPCODE and *decoding*. Step-2 shows how the OPCODE instruction is executed for an example of LUI ( Load Upper Immediate).

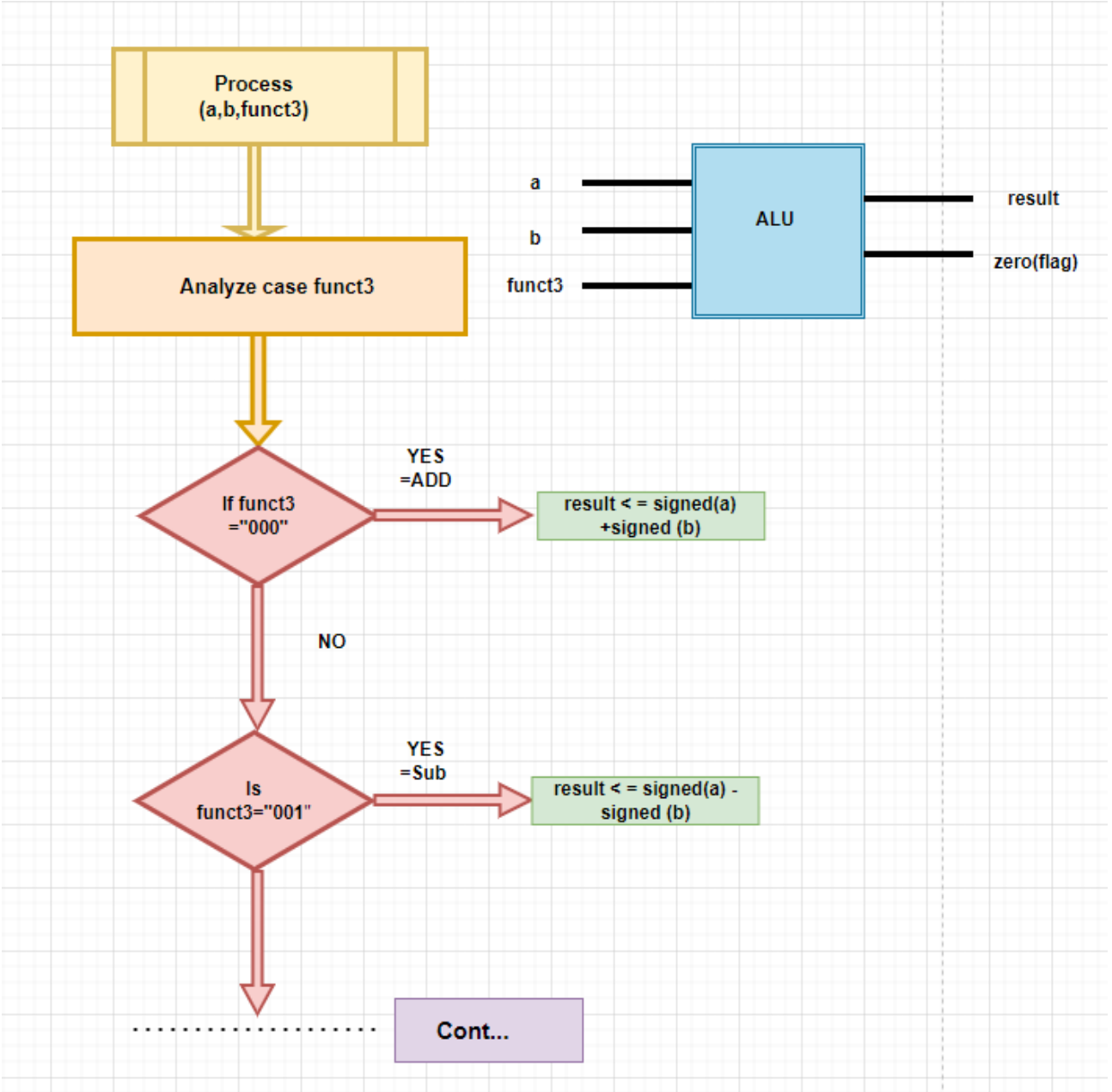

**Fig. 7**: Flowchart showing the operation of ALU

Fig.7 shows the operation of ALU. The three inputs include ‘*a*’, ‘*b*’ and ‘*funct3*’. The value of *funct3* determines the operation performed by the ALU unit and the corresponding output includes *‘result’* and *‘zero(flag)’*.

### E. *Instruction Memory and Data Memory*

The instruction memory contains the program that is to be executed and is 32 bits in length. The address of the instruction memory begins at 0 x 01000000. The data memory is also 32 bits in length and contains the data to be executed. The address should begin at 0 x 80000000. The data memory is accessed by the “load word”(LW) and “store word” (SW) instructions.

### F. *Steps of the RISC Processor operation*

Fig.8 shows the interconnection between the different components in the RISC. The schematic was generated using Vivado 2023.1. The PC value is the address of the next instruction to be fetched from the instruction memory. The fetched instruction could be in the R-format as shown in Fig. 9. The OPCODE ( bit 6 to bit 0) and function field bits (bit 14 to bit 12) are sent to FSM of the Control Unit to decode the instructions. The control unit will then execute its function accordingly as shown through an example of LUI in the flowchart of Fig. 6. The ALU also executes the function according to the signal received from control unit by analyzing “*funct3*”.In the final step the result from ALU is transferred to the Register file, *rd* ( whereas *‘a’* and *‘b’* in ALU are *rs1* and *rs2* accordingly).

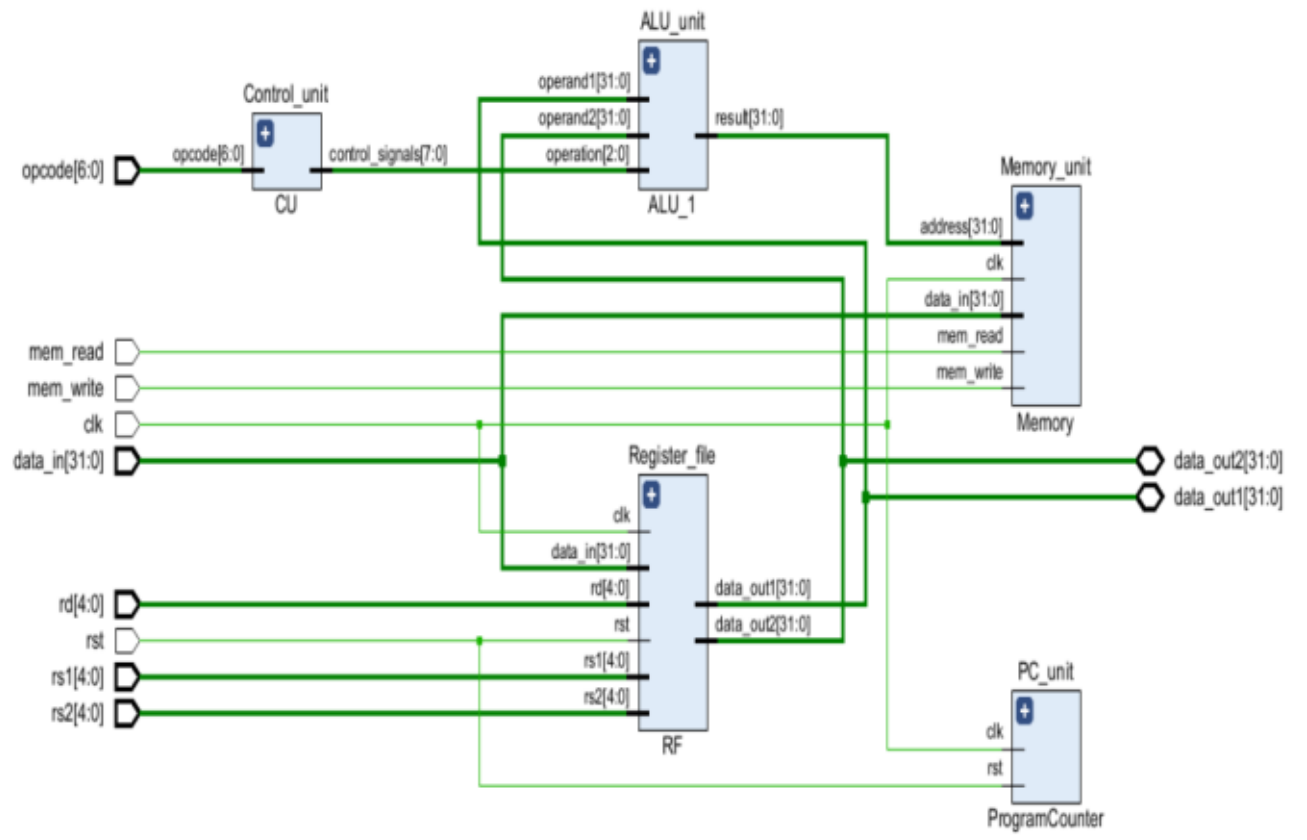

**Fig.8 :** The interconnection between the different components in RISC. Schematic of Elaborated Design RTL Analysis using Vivado 2023.1

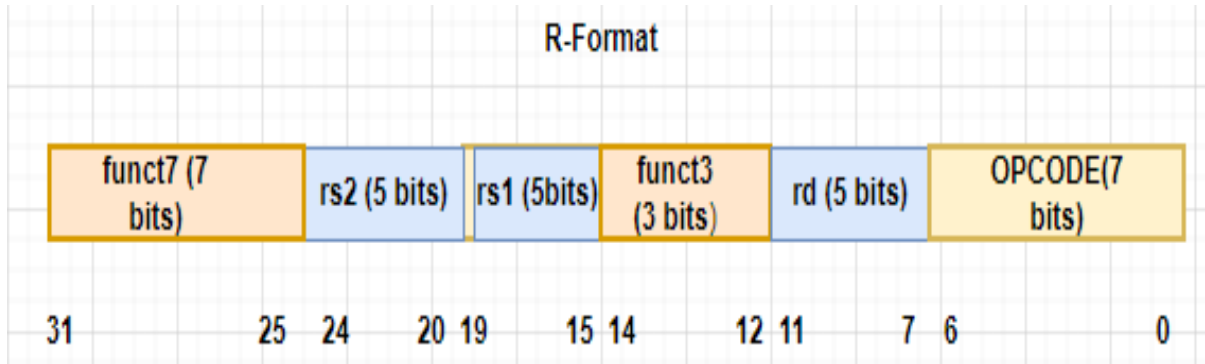

**Fig.9** : R-Format used for instruction in RISC

## IV. Results and Discussion

### A. *Program Counter Testbenches*

Fig. 10 shows the snippet of the timing waveform that was used to verify the testbench operations of the PC. The following *test cases* were verified: (1) the initial value after reset. (2) whether it can increment the PC counter's value . (3) whether the reset is working ( it checks for different clock cycles) (4) the PC increment counter. The timing waveform simulation was compiled, and the following was obtained as shown in Fig.10. At 30 ns, the PC counter increments by 4 that is changing from 40000000 (Hexadecimal which is binary 01000000000000000000000000000000) to 40000004. This shows that the Test-2 is working. A similar increment is again seen at 70 ns and is not shown in the diagram ( to verify test case 4). It is also seen at 40 ns clock cycle, rising edge with the reset-'1/ high', the *pc_out* value again resets to (Hexadecimal which is binary 01000000000000000000000000000000) which was the default value.

Fig. 11 shows the Xilinx Artix-7 FPGA (xc7A35T-1CPG236c) Board that was used to implement and test the PC functionality . V17 was the reset pin that was assigned in the constraint file and when it was set to '1' the PC was returned to *0 x01000000*

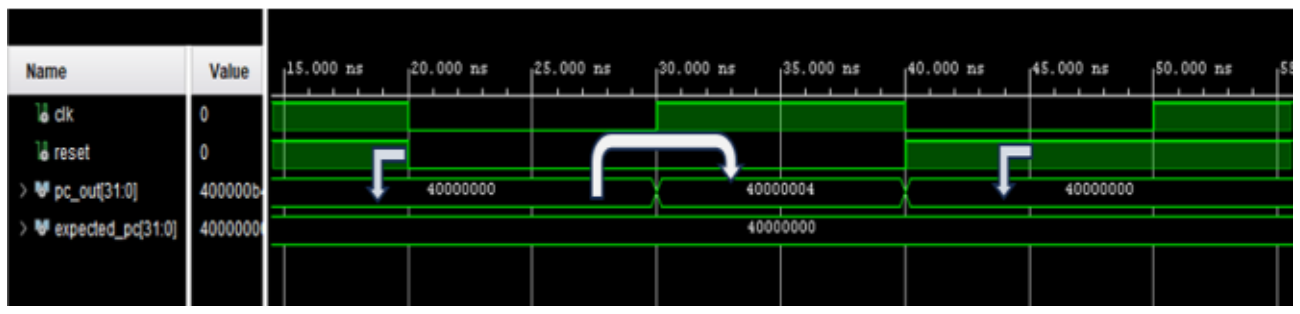

**Fig. 10 :** Timing waveform for the testbenches of Program Counter (PC)

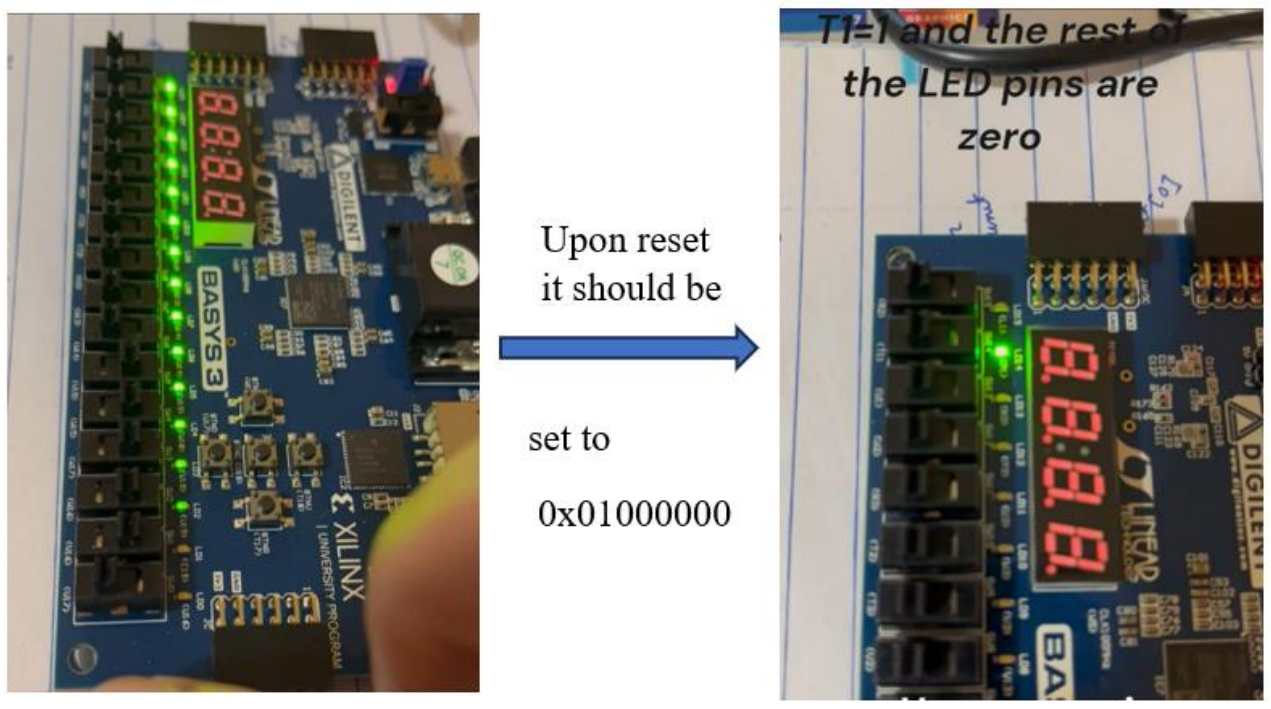

**Fig. 11 :** Basys 3 FPGA Board was used to implement and test the PC functionality**.**

## *B. Register File Testbenches*

The following *test cases* were verified: (1) Checks the initial value after reset. (2) It checks whether it can load address value. (3) It checks whether the reset is working (4) Checks the output of the register. Fig. 12 shows the snippet of the timing waveform that was used to verify the test bench operations of the Register files . At 200 ns, the "*register_write_ enable*" turns *high*, and the register *"write_data"* starts loading, which is 4000 0004 ( in hexadecimal) , and at 100ns, the "*register_read_address_1*" and "*register_read_address_2*" starts to load the data, 01 and 02 respectively . For the "*register_read_data_1*", it's initially at *0000 0000*, until the next clock period upon which it reads the corresponding write data, which is *4000 0004***.**

Xilinx Artix-7 FPGA (xc7A35T-1CPG236c) board was used to program the hardware configuration of the register file. Instead of the 32 bits only 8 bits were used to program the read and write functionalities along with the *reset* and *write_enable* were included in the constraint file. Fig. 13 shows that when *'reset'* pin was set to '1', the write register file was set to "*00000000*". In case *II*, when reset= " 0 ", *write_enable*= "1" ( activated to write), the read registers read the output(LED off = '0' and LED on = '1') . Similarly, for the input *write_data*, switch off= ' 0' and switch on= '1'. The *read registers* read the same output as the *write registers* input as shown in Fig. 13.

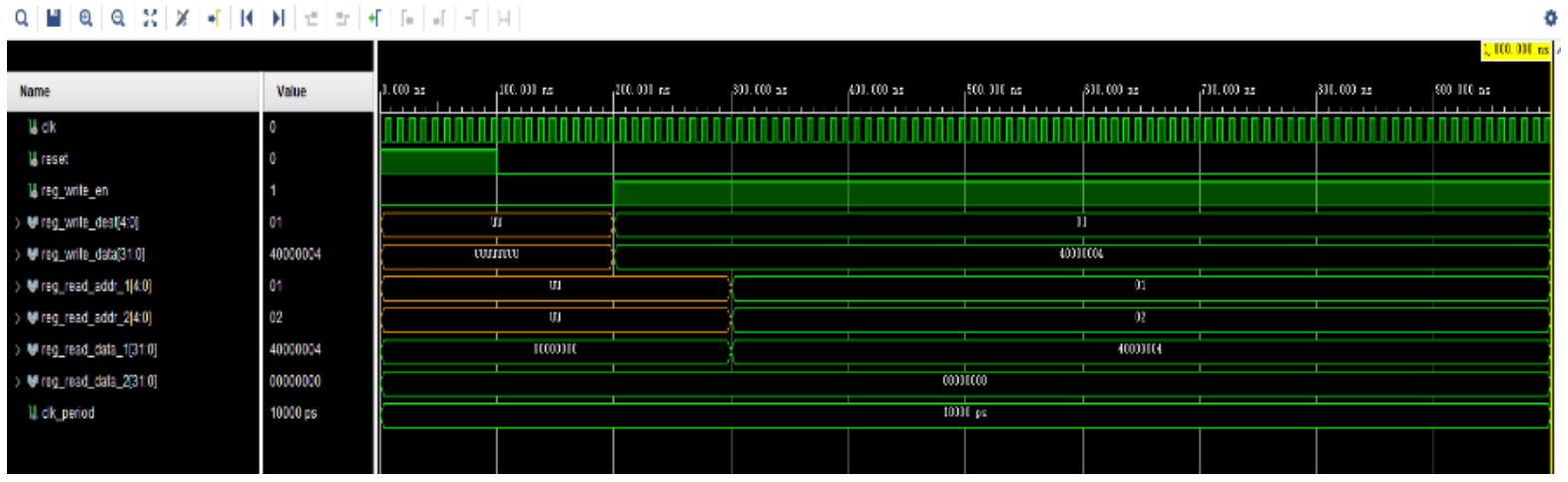

**Fig. 12 :** Timing waveform for the testbenches of Register File.

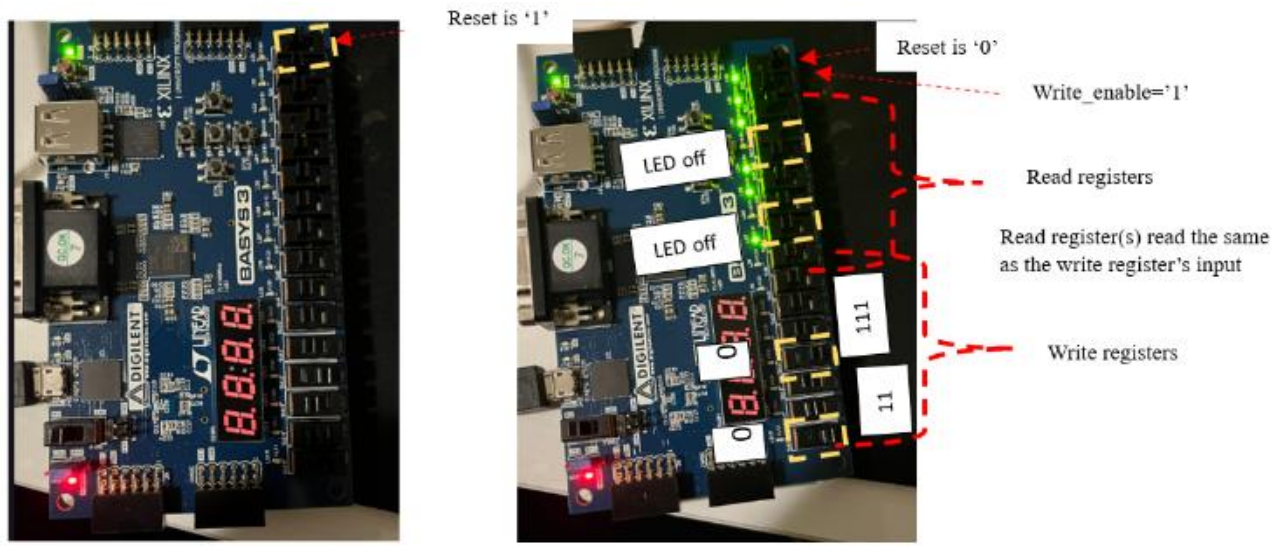

**Fig. 13 :** Basys 3 FPGA Board was used to implement and test the Register File's functionality**.**

## *C. Control Unit Testbenches*

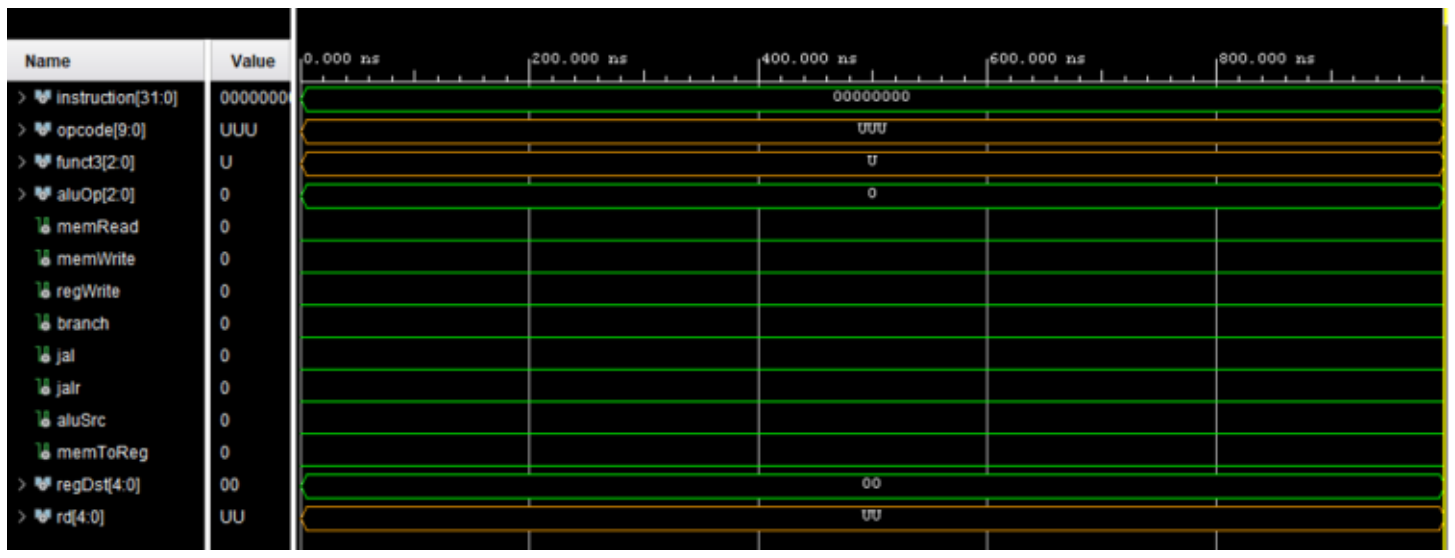

**Fig. 14 :** Timing waveform for the testbenches of Control Unit

Fig. 14 shows the snippet of the timing waveform that was used to verify the testbench operations of the Control Unit. The following *test cases* were checked: (1) Testing *AND* instruction (2) Testing the *load* instruction (3) Testing the *store* instruction (4) Testing *BEQ (Branch if Equal)* instruction (5) Testing *JALR (Jump and Link Register)* instruction (6) Testing *SLTI* (*Set Less than Immediate signed*) instruction (7) Testing *SLTIU* (*Set Less than Immediate-unsigned* )instruction (8) Testing *XORI (XOR with Immediate)* instruction (9) Testing *ORI(OR with Immediate)* instruction and (10) Testing *ANDI(AND with Immediate)* instruction.

The source file for the Control Unit was able to compile along with the corresponding testbenches after a significant number of iterations. The control unit is a core component of the RISC as it contains the FSM that performs the decoding of the control unit.

The integration of the Control unit along with other core components (Register file, ALU, PC, Instruction and Data Memory) in the RV 321 was performed through port map initiation. The timing waveform confirms that the Control Unit was able to function accordingly and is discussed in detail in *Section F of Results and Discussion.*

### *D. Arithmetic Logic Unit (ALU) Testbenches*

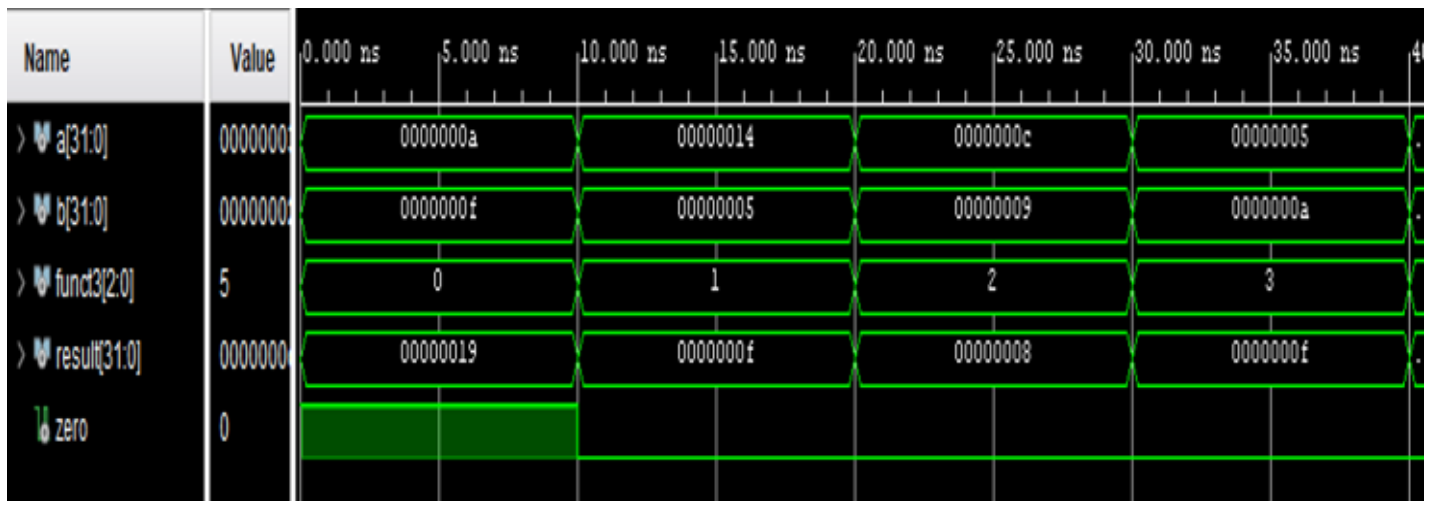

**Fig. 15 :** Timing waveform for the testbenches of ALU

The following *test cases* were verified: (1) Addition (2) Subtraction (3) AND (4) OR (5) XOR (6) shift logic left (SLL) (7) shift logic right (SLR) (8) additional 'add' test (9) additional 'and ' test and (10) additional 'subtraction' test. *Funct3* was used to determine which of the following operations were to be executed.

From the source code of ALU, *funct3=000*:Add( between time frame 0 ns - 10 ns); hence result =(*00000019*) which is addition performed due to the *register a*: 0000000a and *register b:* 0000000f. Similarly, funct3=010:AND ( between time frame 20 ns - 30 ns); hence result =(*0000000f*) which is addition performed due to the *register a*: 00000005 and *register b:* 0000000a.

## *E. Instruction Memory and Data Memory Testbenches*

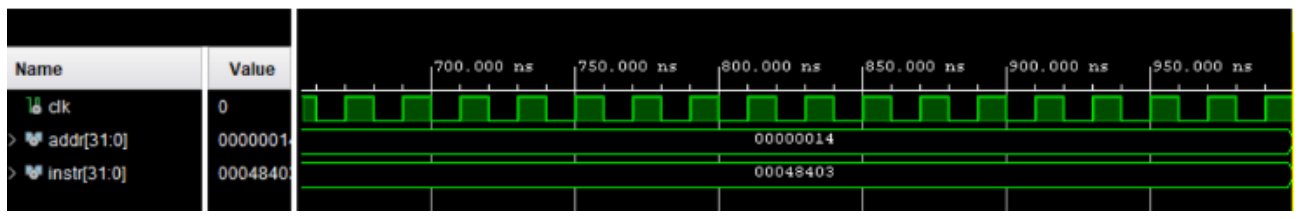

**Fig. 16 (a) :** Timing waveform for the testbenches of Instruction Memory

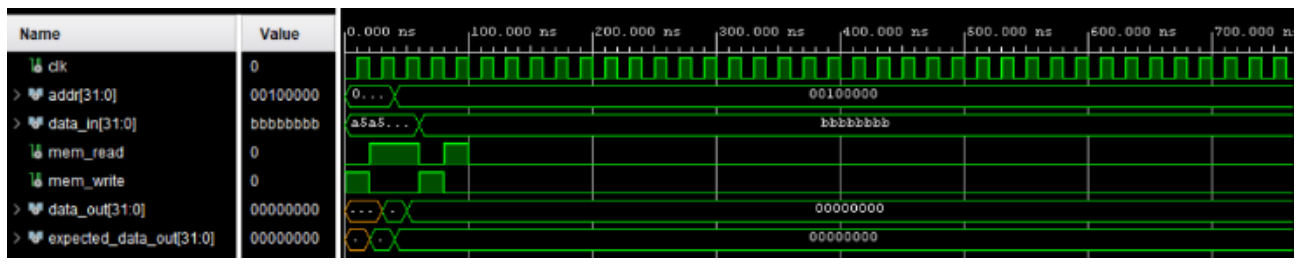

**Fig. 16 (b) :** Timing waveform for the testbenches of Data Memory

The following *test cases* were checked for Instruction Memory as shown in Fig. 16 (a) :

(1) Instruction (according to the R-Format discussed in Fig.9) : "*00000000010000001000000010110111*" . (2) "*00000000010100010000000100010111* (3) *00000000000000010011000011101111* (4) checking a non-initialized instruction and (5) *00000000110001111011111100110011*. From Fig. 16(a), it can be seen from the timing diagram when the *address is :00000014* , the corresponding instruction is *Instr:* 00048403 (Hexadecimal, binary: *00000000010010000100000000011*). This is equivalent to the instruction LW , which translates to: Load 32-bit value at memory address [rs1 value]+(sign extended immediate) and store it at rd.

The following test cases were verified for Data Memory as shown in Fig. 16(b) : (1) writing data (2) reading data (3) read from *read_only memory* (4) writing to *read_only memory*. The fourth test can be verified at around 30 ns, when the *mem_write* is high, add is: x "00100000" and *data_in* is: "BBBBBBBB", the corresponding *data_out=expected_data_out* which in this case is *'00000000'.*

### *F. Testbenches for Combined Processor using Portmap Initiation*

Portmap initiation was used to connect the components listed above. Fig. 17(a) shows the snippet of the timing waveform that was used to verify the testbench operations of the combined processor, RV 321 Processor. The following testcases were checked and are in good alignment with the testbench results. The testbench cases include: (i) R-type instruction for addition. (ii) R-type instruction for subtraction (iii) Memory read (iv) branch instructions (v) memory write (vi) Conditional branch ( if *rs1= =rs2*) (vii) R-type instruction (logical AND) (viii) Memory read(negative offset) (ix) Conditional branch ( if *rs1!=rs2*) (x) memory write(immediate effect) (xi) if *rs1<rs2*) .

Fig. 17(b) shows the testbench for the OPCODE for SB (Store Byte). At 60 ns the opcode is : *"010011*"- This is the opcode for SB ( store byte)- the function of SB is to store the lower 8 bits of *rs* from one register to the other and hence at 60 ns it stores the lower 8 bits of *rs1* to *rs2*.

Fig. 17(c) shows the testbench for the OPCODE for LHU (Load Half Word). At 20 ns the opcode is : "*0000011*"- This is the opcode for LHU (load half word unsigned)- the function of LHU is to Load a 16-bit unsigned value from memory at the address specified by the sum of the contents of register *rs1* and the sign-extended immediate value (*imm*). Zero-extend the 16-bit value to 32 bits and store it in register *rd*. Hence at 20 ns the value of *rd* is updated to "*04 (hexadecimal)*"

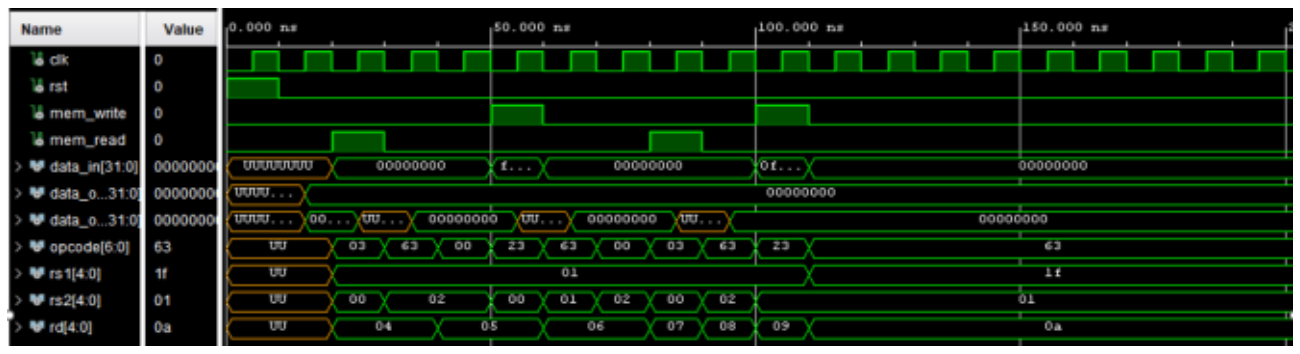

**Fig. 17 (a) :** Timing waveform for the testbenches of Data Memory

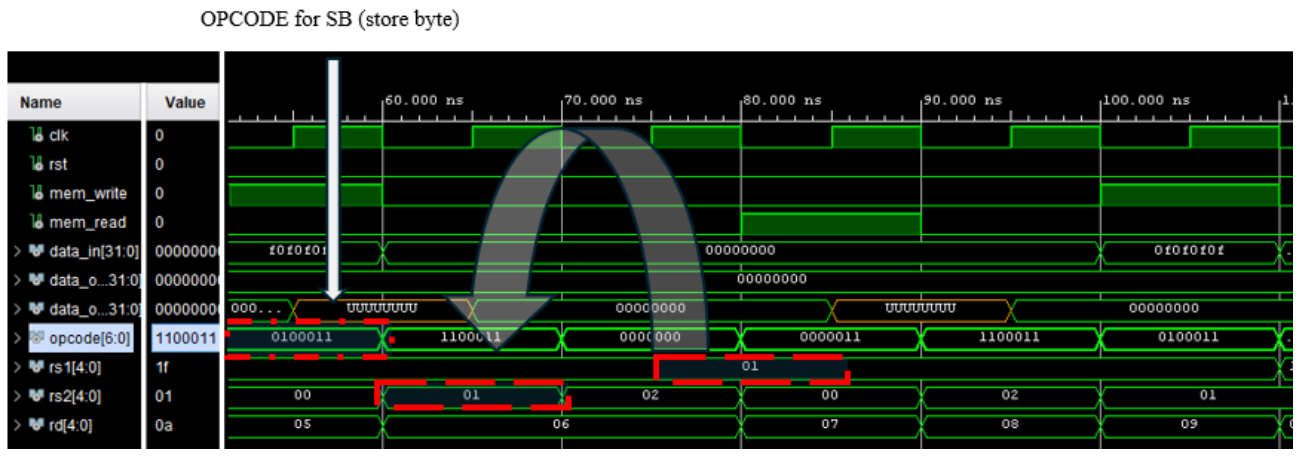

**Fig. 17(b) :** Testbench for the OPCODE for SB (Store Byte)

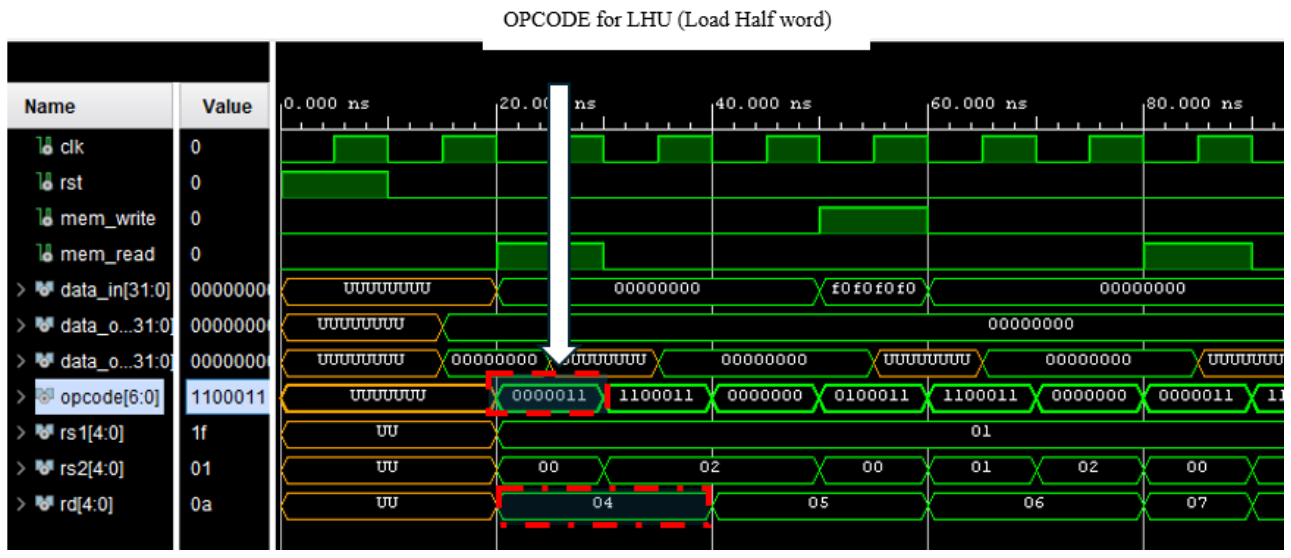

**Fig. 17(c) :** Testbench for the OPCODE for Load Half Word (LHU)

Fig. 20 (a) and (b) shows the specification summary of the power consumption and the synthesized design accordingly . The elaborate design includes 5 cells, 122 Input/Output (I/O) ports and 157 nets. The total chip power consumption is approximately 121 mW and the I/O ports account for the highest power consumption approximately at 74%. This design can be optimized for both (i) power consumption and (ii) execution time and will be focused in our future work.

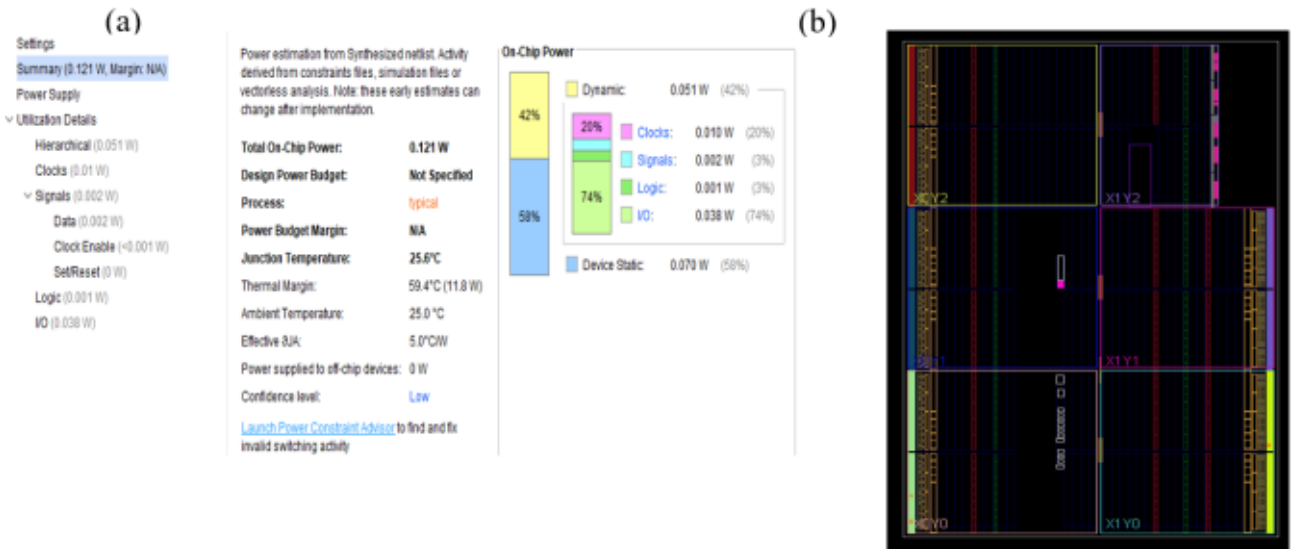

**Fig. 20**: Specification summary of the power consumption and the synthesized design

### *G. Challenges in the design of Processor RV321:*

Table -I compares the four parameters: (i) *Correct output on first iteration* (ii) *Number of errors embedded in the code* (iii) *Number of trials required to achieve the code* and (iv) *Failure to generate the code after three iterations* ; to compare the efficiency of using LLM in programming for this application in RISC. In the design of all the six components there was a similar consensus of failure to generate the accurate VHDL code on the first iteration and in all the cases human intervention was eventually required to fix the bugs. The number of errors and trials required to fix the errors varied and depended mostly on (i) the complexity of the design and (ii) clarity in the prompt. Most of the errors were syntax errors and in certain cases, the generated code was not complicit with the requirement of the system. Several iterations as listed in Table-I to get the desired source file and testbench to compile and execute. Hence it can be concluded for this particular design that though LLM is a great tool for generating a preliminary skeleton for the code, human intervention is eventually required to fix the relevant bugs embedded in the code.

TABLE I

COMPARISON METRICS TO EVALUATE THE PERFORMANCE OF USING LLM AT DESIGNING A RISC

| | **Correct output on initial iteration** | **Number of errors embedded in the code** | **Number of trials required to achieve the correct code** | **Failure to generate the correct code after three iterations** | **Notes** |
|---|---|---|---|---|---|
| PC | 0 | 1 | Greater than 3 | 1 | Human intervention required to correct the code. |
| Register File | 0 | 4 | Greater than 4 | 1 | Human intervention required to correct the code. |
| ALU | 0 | 8 | Greater than 4 | 1 | Human intervention required to correct the code. |
| Control Unit | 0 | 5 | Greater than 10 | 1 | Multiple efforts were required, and senior consultants were used to correct it. |
| Instruction Memory and Data Memory | 0 | 4 | Greater than 3 | 1 | Human intervention required to correct the code. |

## V. CONCLUSION

LLM has been extensively used at generating code-based output. This paper discusses a case study of using LLM to implement VHDL based code to design a RISC. It discusses the different steps associated with the ChatGPT 3.5 LLM at code generation including parsing, tokenization, encoding, attention mechanism, sampling the tokens and iterations (optional). The general architecture of the RISC is also reviewed and the corresponding testbenches are used to verify the design implementation.

A simple proposed model of using metric parameters such as : Correct output on first iteration (ii) Number of errors embedded in the code (iii) Number of trials required to achieve the code and (iv) Failure to generate the code after three iterations; are used to compare the efficiency of code generation. In the design of the components there were significant errors embedded in the generated code and in all cases human intervention was required to fix the bugs.

This paper discusses: (i) the success of using LLM at code generation with appropriate prompts but also realizes that programmers are required to handle bugs (ii) uses a comparison metrics to evaluate the efficiency of code generated by LLM. Future work could include how LLM can be used to (i)"self-correct" its errors- with minimum or no human intervention and (ii)improve the design in terms of both power consumption and timing delays from fetching to execution.